\documentclass[10pt,twocolumn]{article}
\usepackage{graphicx}
\begin{document}

\newtheorem{definition}{Definition}

\title{When Do Differences Matter?  On-Line Feature Extraction
Through Cognitive Economy}

\author{\large David J. Finton\\
\normalsize Laboratory for Molecular and Computational Genomics,\\
\normalsize University of Wisconsin-Madison\\
\normalsize Madison, WI\\
\normalsize Email: finton@cs.wisc.edu}

\maketitle

\begin{abstract}

For an intelligent agent to be truly autonomous, it must be able to adapt
its representation to the requirements of its task as it
interacts with the world.  Most current approaches to on-line feature
extraction are {\em ad hoc;}  in contrast, this paper presents
an algorithm that bases judgments of state compatibility and
state-space abstraction on principled criteria derived from
the psychological principle of {\em cognitive economy.}  The
algorithm incorporates an active form of Q-learning, and partitions
continuous state-spaces by merging and splitting Voronoi regions.
The experiments illustrate a new methodology for testing and
comparing representations by means of learning curves.   Results
from the puck-on-a-hill task demonstrate the algorithm's ability to
learn effective representations, superior to those produced by some
other, well-known, methods.

\end{abstract}

\section{Introduction}

Representation is the foundation for problem-solving:  it provides
the vocabulary and populates the world that we seek to understand
and control.  Although we can sometimes specify a good representation
for particular problems, we have not understood the general learning
problem until we understand how the representation can be learned
along with the behaviors that lead to success in a task.  There
are also important practical reasons for studying autonomous
representation learning.  For example,
success may depend on the agent's ability to learn an effective
representation from scratch if the task is poorly understood, or
if there are too many possible scenarios for us to work out a
complete specification of the state-space in advance.  Even when
it would be possible to design the representation beforehand, it
might not be a cost-effective use of programmer time, especially
if the representation will later need to be updated as the task
environment changes.

This paper presents a new approach to the problem of autonomous
representation learning,
by applying the psychological principle of cognitive economy
\cite{Rosch78} to the domain of reinforcement learning
\cite{KaelblingLittmanMoore96,SuttonBarto98}.

\subsection{Effective representations}

One of the hard problems facing any theory of cognition is that of
finding principled ways of specifying when states of the world are
``the same'' and when they must be distinguished.  Distinguishing
every possible state of the world from every other state makes
learning intractable except in very small, discrete state-spaces;
but the agent cannot learn the task if
the representation groups together states of the world that
require different behaviors.
Ideally, the agent should learn which states must
be distinguished, while avoiding irrelevant distinctions that
prevent it from generalizing its learning over states that are
``the same kind of thing'' in its task.  How can an agent learn
such a representation without knowing about the task beforehand?

\subsection{Function approximation and value prediction}

The typical approach to reinforcement learning represents the agent's
knowledge of the world in terms of an action-value function, $Q(s, a).$
This function gives the long-term estimate of reward that results from
taking action $a$ from state $s,$ and following a greedy policy thereafter
(that is, the agent chooses the action with the highest value in all
subsequent states) \cite{WatkinsDayan92,SuttonBarto98}.
The original descriptions of Q-learning assumed a
{\it discrete representation:}  the action-value function was assumed to be
stored as a table having a separate row for the values of each distinct state
\cite{WatkinsDayan92}.  To extend the approach to large
and continuous state-spaces, we may store the values $Q(s, a)$
more compactly as a parameterized function of $s$ and $a;$  the learning
problem then becomes an exercise in function approximation, where the
agent responds to its experiences in the world by adding features or
tuning parameters so that it minimizes the mean-squared error (MSE)
in the value predictions, $Q(s, a).$
Sutton and Barto comment that ``our ultimate purpose is to use the
predictions to aid in finding a better policy.  The best
predictions for that purpose are not necessarily the best for
minimizing the MSE.  However, it is not yet clear what a more useful
alternative goal for value prediction might be,'' \cite[p. 196]{SuttonBarto98}.

This paper presents an alternative goal for value prediction, based on
the insight that some $Q$-value errors will have no effect on the
agent's ability to perform its task.  The agent
learns faster when it can generalize over ``similar'' states: states
that agree on the preferred action and expectation of reward.  Because
similar states may differ on the expected values of non-preferred actions,
grouping these states may increase the overall prediction error---even
though these differences do not impact the agent's performance in the
task.  In contrast, some states should be considered incompatible
because ignoring their differences leads the agent to make bad decisions;
such states must not be grouped together.  This paper
presents principled criteria for deciding when the differences matter
and when they may be ignored.

\subsection{Feature extraction and state abstraction}

Assume a representational model in which the value
function $Q$ is written in terms of a weighted set of feature detectors:
\[
 Q(s, a_j) = \sum_i w_{ij} f_i(s)
\]
This model characterizes much of the work on
function approximation \cite[Ch 8]{SuttonBarto98}.
For example, we can describe a partition representation
by defining a feature $f_i$ for each state-space region
$S_i,$ such that $f_i(s) = 1$ for $s \in S_i$ and $f_i(s) = 0$ for
all all other states $s.$  This same model
encompasses discrete (``look-up table'')
representations, partition representations, tiled representations
(such as CMAC \cite{Albus81}), perceptrons, and radial-basis function
networks, depending on the definition of the $\{f_i\}.$

The features and the state-space regions lead to complementary
ways of looking at the same function approximation process.  We see that
the function $f_i$
is a rule (intension) describing a set $S_i$ of states (extension)
that are grouped together.  We may define the state-space groupings
in terms of the features, or equivalently, we may define the features
in terms of the groupings.  When the features $\{f_i\}$ are
continuous-valued, the corresponding state groupings $\{S_i\}$
will be fuzzy sets.
If we think of the state groupings as determining the feature detectors,
we can think of the $S_i$ as ``generalized states,'' which determine
how action values are stored---just as the individual states do in the
discrete representation.  This
{\it duality of features and state-space grouping} is important because
the grouping of states is the key concept for deciding
which $Q$-value differences
matter in the agent's task.  When we consider function approximation without
discerning the role played by state abstraction, it becomes difficult to
determine how the differences affect the agent's ability to choose
the correct behavior at each state.
 
\subsection{Ad hoc approaches to state abstraction}

One approach to representation is to continue to subdivide
the state-space until its
resolution is adequate to distinguish states that are not ``the
same kind of situation.''  Ideally, the representation will make
finer distinctions in parts of the space where the differences
matter, and simplify the representation of other areas.  In other
words, the representation should have a resolution that varies
throughout the state-space according to the demands of the task.
Function approximation methods that simply cluster the task inputs
cannot provide this kind of representation because they are blind
with respect to the task requirements.  Although we can sometimes specify
important areas of the state-space for particular tasks, it is
hard to do so in a general way.

For example, \cite{HuFellman96} assumed that states closest to
the initial state required the finest resolution;  \cite{Moore91}
assumed that states closest to paths taken by the agent through
the state-space were most important; \cite{Holdaway89} assumed that
the most frequently-seen areas of state-space were most important.
These criteria led to effective representations for the tasks being
studied, but we can readily imagine tasks in which these criteria are
irrelevant.  We need representational criteria that explain why---and
when---these strategies identify state-space differences that are relevant
to the agent's task.  The key is to define important differences in
terms of more general criteria for representational adequacy.

\subsection{Cognitive economy}

Cognitive economy generally refers to the combined simplicity and
relevance of a categorization scheme or representation. Natural
intelligences appear to adopt categorizations with high cognitive
economy in order to make sense of the sea of stimuli impinging on
their senses without overloading their bounded cognitive
resources.  Under the heading
{\em Cognitive Economy}, Eleanor Rosch writes of the
``common-sense notion'' that the function of categorization is to
``provide maximum information with the least cognitive effort,''
``conserving finite resources as much as possible'' \cite[p. 28]{Rosch78}.
Then she writes (p. 29):
\begin{quote}
\ldots one purpose of categorization is to reduce the
infinite differences among stimuli to behaviorally and cognitively
usable proportions. It is to the organism's advantage not to
differentiate one stimulus from others when that differentiation
is irrelevant to the purposes at hand.
\end{quote}
Cognitive economy results when the representation makes
task-relevant distinctions while ignoring irrelevant information.  This
form of  selective generalization presents the agent
with a simpler working environment for its task.  To apply this
principle to reinforcement learning, we must define criteria for
{\em relevant distinctions} without appealing to any task-specific
information.  This paper defines relevant distinctions in terms
of the amount of reward that the agent stands to lose by ignoring
them.  The resulting criteria characterize state-space distinctions
that are important for the agent to maximize its reward in the task.
In this way, the task's reward function determines relevance in the
agent's world.

\section{Representational criteria}

Given a pair of states, $s_1$ and $s_2,$ we want to know
whether the agent may safely group them together so that they
share action values---will this state generalization
cause the agent to lose reward?  One way to find out is to
consider the two states separately, comparing the
expectation of reward for actions taken from $s_1$ with the
expectations from $s_2.$  For example, we can consider the sum of
the immediate reward given for taking action $a$ from $s_1$ and
the value of the resulting state.  In this way, we push our dependence
on the value predictions one step into the future, and bypass any
$Q$-value error caused by inappropriate generalization of $s_1.$  We may
compare these ``look-ahead'' values with $Q(s_1, a)$ in order to
decide whether representational errors are compromising the agent's
ability to choose actions that maximize its reward in the task.  We
may also compare the look-ahead values for $s_1$ and $s_2$ in order
to judge whether they may safely be included in the same
``generalized state.''  If not, the agent should refine its
representation so that it learns their action values separately.  

The next section presents a vocabulary for this discussion.  Specifically,
it defines value functions and action preference sets to be used in
criteria for representational adequacy and state compatibility.

\subsection{Preference and value functions}

\begin{definition}[Action value]
    \[ Q(s, a_j) = \sum_i w_{ij} f_i(s) \]
\end{definition}
This is the same definition given earlier, restated here for convenience.

\begin{definition}[State value]
    \[ V(s) = \max_j Q(s, a_j) \]
\end{definition}
Thus $V(s)$ represents the long-term reward given by the best
action available from $s.$  This is the standard definition of state value,
given by \cite{WatkinsDayan92,SuttonBarto98}.

\begin{definition}[Preferred action set]
\[
    \mbox{pref}_{\epsilon}(s) = \{ a_j  : Q(s, a_j) \geq V(s) - \epsilon \}
\]
\end{definition}
The preferred action set contains the action or actions that appear
to maximize the agent's expected reward.  Thus $\mbox{pref}_0(s)$
gives the action(s) with value $V(s),$ while taking 
$\epsilon > 0$ makes the selection less stringent, and causes the
preference set to contain all actions with value within $\epsilon$ of $V(s).$

The functions for our one-step look-ahead are analogous to the
definitions of $Q,$ $V,$ and $\mbox{pref}_{\epsilon}(s).$

\begin{definition}[Look-ahead action value, case 1]
\label{def:q1}
Suppose that taking action $a$ from state $s$ always results in the following
transition:
\[
    s \stackrel{a}{\longrightarrow} r, s'
\]
Then we define
\[
    Q1(s, a) = r + \gamma V(s')
\]
\end{definition}
Like the function $Q(s, a),$ $Q1(s, a)$ represents the expected
discounted future reward when the agent chooses action $a$
from state $s.$  The parameter $\gamma$ is the Q-learning discount
for future rewards \cite{WatkinsDayan92,SuttonBarto98}.  This definition
holds when the task rewards and state transitions are deterministic.

\begin{definition}[Look-ahead action value, case 2]
    \[
        Q1(s, a) = \sum_{s'} {\cal P}^a_{s s'} [ {\cal R}^a_{s s'}
                                                + \gamma V(s') ]
    \]
\end{definition}
In many tasks the rewards and state transitions will either be stochastic,
or may appear stochastic simply because of imperfect function approximation.
In this case, we need to modify Definition~\ref{def:q1}: we replace the
immediate reward $r$ with its expected
value, ${\cal R}^a_{s s'},$ and we consider all possible resulting states $s',$
weighted by their probability of occurrence, ${\cal P}^a_{s s'}.$ 

Now we may define look-ahead versions of the state value and preference functions by replacing use of the value function $Q$ with $Q1$.

\begin{definition}[Look-ahead state value]
     \[ V1(s) = \max_j Q1(s, a_j) \]
\end{definition}
\begin{definition}[Look-ahead preferred action set]
     \[
        \mbox{pref1}_\epsilon(s) = 
        \{ a_j : Q1(s, a_j) \geq V1(s) - \epsilon \}
     \]
\end{definition}

\subsection{Representational adequacy}

\begin{definition}[$\epsilon$-adequacy]
    \label{3def:adequacy}
    Let $\delta$ be given, and assume that the agent always
    selects an action from $\mbox{pref}_\delta(s).$
    We will say that a representation of the state-space
    is an $\epsilon$-adequate representation for
    $\delta \leq \epsilon,$ if for every state
    $s$ reachable by the agent, the following
    two properties hold:
    \begin{equation}
        \mbox{pref}_\delta(s) 
                 \subseteq \mbox{pref1}_\epsilon(s) \label{rule:policies}
    \end{equation}
    \begin{equation}
        | V1(s) - V(s) | \leq \epsilon \label{rule:values}
    \end{equation}
\end{definition}

Meeting the $\epsilon$-adequacy criterion guarantees that the state
generalization at $s$ does not prevent the agent from being able
to learn the correct policy at $s$ or mislead the agent at an
earlier state as to the desirability of $s.$  Thus, this criterion
defines a standard for representational accuracy at individual
states, guaranteeing that the harmful effects of state generalization
are kept in check and that the agent can learn to make sound
decisions.

This standard defines an {\em adequate representation} as one
that makes the distinctions needed for the task to remain learnable.
It characterizes relevant distinctions in terms of the amount
of reward that the agent stands to lose by ignoring them.  This is
the approach taken in \cite{Finton02}, which introduces the {\em
incremental regret} of a representation at time $t$---the amount
of reward the agent loses when it groups its current state, $s_t,$
in some category $S.$  A perfect representation would have an
incremental regret of 0 at each step, because the representation
would allow the agent to learn to distinguish the best action for
every state.

Non-zero incremental regret arises from two kinds
of representational error: grouping states that have different
policies, and grouping states that have very different values.
First, if the representation groups $s_t$ with the wrong states, the
action that appears best for the group, $S,$ may be sub-optimal for
$s_t.$  This could lead the agent to take the wrong action from $s_t.$
Second, if the value of $s_t$ is very different from the value
of other states in $S,$ the agent might not recognize
that arriving in $s_t$ is a special opportunity (or pitfall), because
the action values are averaged over all the states in $S$---not just $s_t.$
This could cause the agent to
make the wrong choice from $s_{t-1}.$  In both cases, the agent
makes wrong decisions, resulting in lost reward in the task.
The $\epsilon$-adequacy criterion limits the amount of lost reward
by comparing the policy and value predicted by function approximation
with the results of a one-step look-ahead.  Thus
the actions that appear to have the best value for $S$ must be
good actions for $s_t$ (Equation~\ref{rule:policies}, concerned with
{\em policy distinctions}), and
the state value of $s_t$ must be close to that
predicted by the information given for the category $S,$
(Equation~\ref{rule:values}, for {\em value distinctions}).
This bounds the incremental regret at $s_t$ by $\epsilon.$

The $\epsilon$-adequacy criterion allows us to take a representation for
which the action values are known, and test whether it makes the
state-space distinctions that are important for a particular task.  If the
action values are still being learned, such judgments are only provisional.
It is useful to have an additional criterion for {\em state compatibility,}
if the agent is to learn its representation along with the action values.

\subsection{State compatibility}

Although the $\epsilon$-adequacy criterion provides an objective
standard for an adequate representation---one which allows the
agent to learn its task---these characteristics of the representation
are really the outcome of the particular distinctions the representation
makes or fails to make between individual states.
In practice, it is often more useful to be able to evaluate the compatibility
of two states than the compatibility of a state with a region, because
a poorly-chosen region could be incompatible with {\em all} its member
states.  This can also happen with good regions that simply have not
had their action values updated for a long while.

Thus we need to
bridge the gap between the high-level description of adequate
representations and the low-level decisions the agent must make as
to which states must be kept separate.  In other words, when must
the representation distinguish states and when may it generalize
over states, in order for it to be $\epsilon$-adequate?

\begin{definition}[State compatibility]
\label{3def:compatibility}
Let $\delta$ be given, and assume that the agent always
selects an action from $\mbox{pref}_\delta(s).$  Assume
that our goal is to produce an $\epsilon$-adequate
representation, where $\epsilon \geq \delta.$

We will say that states $s_1$ and $s_2$ are {\em compatible} in case
the following three conditions hold:
\begin{equation}
    \mbox{pref1}_\epsilon(s_1) = \mbox{pref1}_\epsilon(s_2)
    \label{3compat-prefs}
\end{equation}
\begin{equation}
    | V1(s_1) - V1(s_2) | \leq \delta \label{3compat-vals}
\end{equation}
and
\begin{eqnarray}
    \mbox{pref1}_{\delta}(s_1) = \mbox{pref1}_{\delta}(s_2)
                             & \mbox{ if } \delta \leq \epsilon / 2 
                                \label{3compat-consist1}  \\
    \mbox{pref1}_0(s_1) = \mbox{pref1}_0(s_2)
                             & \mbox{ otherwise }
                                 \label{3compat-consist2}
\end{eqnarray}
\end{definition}

The criteria consist of three rules.  The first rule ensures that
the same actions appear desirable in each state.  The second rule
requires that the values of the states are close, based on a one-step
look-ahead.  The purpose of the third rule
(Equations~\ref{3compat-consist1} and \ref{3compat-consist2}) is to
ensure that the action which appears to be the best for a set of
compatible states is, in fact, a pretty good action for any of the
states in the set.  This is difficult to guarantee when the
compatibility criteria are written for pairs of states,
rather than in terms of the whole set.  That is why the criteria
demand equality of the preference sets instead of merely requiring
the preference sets to overlap.
When $\delta \leq \epsilon / 2,$ the looser restriction of
Equation~\ref{3compat-consist1} allows the states to have slightly
different values for the top actions, making the criteria more
suitable for a practical algorithm which must account for real-world
noise in the value estimates.  The cut-off value of $\epsilon / 2$
appears to come from the sum of the errors allowed by combining
Equation~\ref{3compat-vals} with Equation~\ref{3compat-consist1}.

These issues
are worked out in \cite{Finton02}, which also offers
a proof that for partition representations, separating incompatible states
according to Definition~\ref{3def:compatibility} guarantees
$\epsilon$-adequacy of the representation.  Definition~\ref{3def:compatibility}
thus describes criteria which are sufficient to produce
$\epsilon$-adequate representations.  The task of finding a set
of {\em necessary} and sufficient conditions remains future work.

The representational criteria thus allow the system to detect relevant
distinctions while generalizing over similar states.  These criteria
express the principle of cognitive economy in terms of representational
adequacy and state compatibility.  Since the criteria do this by
examining the values of the actions available to the agent in its task,
they allow feature extraction to proceed without depending on any other
task-specific knowledge.  In this sense, the approach is a principled
one, and a solution to the general problem of representation learning by
autonomous agents.

\section{An Algorithm for On-Line Feature Extraction}

The representational criteria not only provide a basis for understanding
how accurately the action values must be learned, but these criteria
also provide the means for analyzing and improving representations for a
particular task.  The most challenging application of these ideas is to
learn the representation along with the rest of the task, especially
when the system is forced to start from scratch, regarding the task
environment as a black box.  Here success depends on the integration of
representation-learning with the rest of the system.  In particular,
learning the action values requires an adequate representation, yet the
representational criteria depend on the accuracy of the
(partially-learned) action values.  Furthermore, changes made to the
representation may cause additional changes in the action values.  This
section presents an online system that meets these challenges as it
learns its representation along with the rest of the task.  Although
this system is just one possible implementation of the ideas, its
success is an argument for the utility and robustness of the
representational criteria.

The algorithm combines Q-learning \cite{WatkinsDayan92} with an active
strategy for remembering ``surprising'' states and examining them at the
ends of trials.  When the system's current action leads to unexpected
results, it pushes the current state on a {\em replacing-stack} data
structure.  (If the state belongs to the same region as an earlier state
on the stack, the earlier state is removed).  At the ends of trials, the
system conducts mini-trials from the states on its stack, investigating
the most recent surprising states first.  These investigations produce
action-value profiles which the system uses to update its action values,
and also to determine whether the representation adequately represents
the surprising states.  If these states are not compatible with the
prototype states for their regions, the system adjusts the state-space
representation accordingly.  The key differences from Q-learning are the
use of the $\epsilon$-adequacy criterion
(Definition~\ref{3def:adequacy}) to detect surprising states, and use of
the state compatibility criterion (Definition~\ref{3def:compatibility})
to decide when to separate two states.  In addition, this version of the
algorithm sometimes selects starting states for its experiments on the
basis of its stack, instead of always beginning at the same ``start''
state and proceeding to a terminal state.

The state-abstraction section of the algorithm is built upon a
nearest-neighbor representation of the state-space.  The partition
regions are the Voronoi regions about each of a series of prototype
states given to the representation.  (A Voronoi region is the set
of points closer to a particular prototype than to the other
prototypes). Some regions consist of a single prototype and its
Voronoi region, while others are compound regions consisting of a
set of merged Voronoi regions.  The compound regions are represented
by a primary prototype state; this state is taken as the representative
state for any of the states which fall in that region, even though
some other state may be their nearest-neighbor prototype.  If the
state to be classified lies within a simple, un-merged region, its
primary prototype will be the nearest-neighbor.  When a state is
judged incompatible with the prototype state for its region, we
split the region by simply adding the surprising state as a new
prototype; it then becomes the primary prototype of a new Voronoi
region in the space.

\subsection{The top level of the algorithm}

The top level of the algorithm is given in Figure~\ref{fig:algtop}.  It
outlines the function {\tt get\_action,} which is called with the results
of the transition
\[
    s_{t-1} \stackrel{a}{\longrightarrow} r, s_t
\]
\begin{figure}[h]
\centering
\fbox{
\begin{minipage}[t]{\linewidth}
   \begin{tabbing}
xxxx \= xxxx \= xxxx \= xxxx \= xxxx \= xxxx \= \kill

      $j \leftarrow$ region($s_{t-1}$) \\
      $k \leftarrow$ region($s_t$) \\
     if terminal($s_t$) or {\tt reliable\_source}($k$) then \\
        \> if terminal($s_t$) then\\
        \>    \> $Q_{\mbox{new}} \leftarrow r$ \\
        \> else \\
        \>    \> $Q_{\mbox{new}} \leftarrow r + \gamma \max_i Q(k, i)$ \\
        \> $Q(j, a) \leftarrow (1 - \alpha) Q(j, a) + \alpha Q_{\mbox{new}}$  \\
        \> $\mbox{Updates}(j, a) \leftarrow \mbox{Updates}(j, a) + 1$ \\
\\
\pushtabs
    if \= not $\epsilon$-adequate($s_{t-1}$) \\
        \> or $j$ has never been investigated \\
        \> or a weighted coin flip returns heads \\
     then \= push ($s_{t-1}, j$)
             on the stack \\
\poptabs
\\
if terminal($s_t$) then \\
    \> {\tt process\_stack()} \\
else \\
    \> return {\tt next\_action()}
   \end{tabbing}
\end{minipage}
}
   \caption{\tt get\_action($s_{t-1}, a, r, s_t$)}
   \label{fig:algtop}
\end{figure}
This function updates the action values and then calls a function
to select the next action.  The action
value update is only applied when the new state,
$s_t,$ is considered reliable:
the function ${\tt reliable\_source(k)}$ is true when 
$\mbox{Updates}(k, b) \geq$ MIN\_UPDATES for some
action $b.$  Rather than looking at the number of visits, this criterion
determines whether any of the action values for region $k$ have
been updated a certain minimum number of times.  The value of
MIN\_UPDATES
was $3$ for the experiments reported here.  Checking for experience
in this way is an enhancement that could be applied to any
value iteration algorithm, and not essential to the algorithm.

The $\epsilon$-adequacy test done here to detect surprising states
is a simplification of the
one defined in Definition~\ref{3def:adequacy};  it uses the simpler,
non-stochastic definition of $Q1$ (Definition~\ref{def:q1}) to compare
the recently experienced transition with the action-value profile
$Q(j, \cdot).$ 

Typically, a driver process invokes {\tt get\_action} to begin a trial
from some start state, $s_0.$  During a trial, the algorithm updates
action values as Q-learning would, and then chooses the next
action according to its current policy and action values.  The
algorithm continues to process state transitions fed to it by the driver,
but may also initiate active exploration at the end of trials by
invoking {\tt process\_stack()}.

\subsection{Active state investigations}

\begin{figure}[htb]
\centering
\fbox{
\begin{minipage}[t]{\linewidth}
\begin{tabbing}
xxx \= xxx \= xxx \= xxx \= xxx \= xxx \= \kill
 while not empty( replacing-stack ) \\
     \> $s \leftarrow$ pop( replacing-stack ) \\
     \> $j \leftarrow$ region($s$) \\
     \> if FE\_TIMER() then \\
     \>    \> $\mbox{\tt update\_representation}(s, j)$ \\
     \> else \\
     \>    \> \mbox{\tt investigate}($s$) \\
\pushtabs
     \>    \> for each $a$: if reliable($a$) then \\
    \>     \>     \>    $\mbox{Updates}(j, a) = \mbox{Updates}(j, a) + 1$ \\
    \>     \>     \>    $Q(j, a) \leftarrow$ \=$Q(j, a)$ \\
    \>     \>     \>    \mbox{    } $+ \mbox{\tt alpha}(j, a) [Q1(s, a) - Q(j, a)]$ \\
\poptabs
\end{tabbing}
\end{minipage}
}
\caption{\tt process\_stack()}
\label{fig:processstack}
\end{figure}
Figure~\ref{fig:processstack} outlines the function {\tt process\_stack()},
which investigates states that appeared ``surprising.''  Because the
stack is a last-in, first-out data structure, {\tt process\_stack()}
explores states that occur at the ends of episodes before it
explores earlier states.  This property allows the system to focus
on the frontier between unlearned states and states whose
action values have already been grounded in the reward given by
the environment.  Actions leading to terminal states are learned
first, then the action values of states one step earlier. As the
system learns about states near the ends of trials, they stop being
surprising, and the system focuses its attention on states which
precede those states.  In this way, the action values are learned
from the end states backwards to the beginning states, but without
all the extra action-value backups from internal states whose values have
not yet been learned.

Because the stack is implemented as a {\em replacing-stack,} pushing any
item causes the stack to remove any previous occurrence of that item
before adding the new one.  To enable the system to cope with
continuous state-spaces, in which the same exact state might never
be repeated, the stack regards states from the same region as ``the
same.''  Therefore pushing a state removes any other states having
the same region from the stack.  This ensures that the stack size
does not grow without bound:  the number of items on the stack is
limited by the number of state-space regions in the representation,
and a particular region will not be explored more than once for
any session.  These are desirable qualities for cyclic tasks like
the puck-on-a-hill task, because a single region might
otherwise fill the stack with states seen during repeated passes
through that region.

\begin{figure}[htb]
\centering
\fbox{
\begin{minipage}[t]{\linewidth}
\begin{tabbing}
xxx \= xxx \= xxx \= xxx \= xxx \= xxx \= \kill
for each action $a$ \\
    \> $\hat{s} \leftarrow s$ \\
    \> $\hat{r} \leftarrow 0$ \\
    \> $\mbox{steps} = 0; \; \mbox{discount} = 1.0$ \\
\\
\pushtabs
    \> repeat \\
    \>    \> $\hat{s} \stackrel{a}{\longrightarrow} r, s'$ \\
    \>    \> $k \leftarrow \mbox{region}(s')$ \\
    \>    \> $\hat{r} \leftarrow \hat{r} + \mbox{discount} * r$ \\
    \>    \> $\mbox{discount}  = \gamma * \mbox{discount} $ \\
    \>    \> steps $\leftarrow$ steps $+ 1$ \\
    \>    \> $\hat{s} \leftarrow s'$ \\
    \> until \=$\mbox{\rm region}(s') \neq \mbox{region}(s)$ or \\
    \>         \>$\mbox{steps} \geq \mbox{MAX\_STEPS}$  or \\
    \>         \> terminal($s'$) \\
\poptabs
\\
    \> $\mbox{reliable}(a) \leftarrow \mbox{\tt reliable\_source}(k)$ \\
    \> if terminal($s'$) then \\
    \>    \> $Q1(s, a) \leftarrow \hat{r}$ \\
    \> else \\
    \>    \> $Q1(s, a) \leftarrow \hat{r} + \gamma \max_i Q(k, i)$ \\
   \end{tabbing}
\end{minipage}
}
\caption{\tt investigate($s$)}
\label{fig:investigate}
\end{figure}
The system investigates a state by conducting a mini-trial for
each possible action from that state (Figure~\ref{fig:investigate}).
These trials only last long
enough for the agent's state to enter another region or for it to reach
a terminal state.  In the case of an action that leads back to the
same region, the investigation times out after a certain number of
steps.  If a single action takes the agent out of the region, the mini-trial
stops after one action.

Feature extraction is not performed after every trial.  If this
investigation is not one in which the system will perform feature
extraction, the system applies the results of its investigation by
updating action values according to the new profile.  Like the
Q-learning updates done in the top level of the algorithm, these
updates only back up values when the resulting states (from the
investigations, in this case) are determined to be reliable.  Unlike
those updates, the learning rate for the active updates decreases
with the number of times the action value has been updated, until
it hits a specified minimum value (Figure~\ref{fig:alpha}).
\begin{figure}[htb]
\centering
\fbox{
\begin{minipage}[t]{\linewidth}
\begin{tabbing}
xxx \= xxx \= xxx \= xxx \= xxx \= xxx \= \kill
 if $\mbox{Updates}(j, a) \leq \mbox{ENOUGH\_SAMPLES}$ \\
    \> return $1.0 / \mbox{Updates}(j, a)$ \\
 else \\
    \> return $1.0 /  \mbox{ENOUGH\_SAMPLES}$
   \end{tabbing}
\end{minipage}
}
\caption{\tt alpha($j, a$)}
\label{fig:alpha}
\end{figure}

\subsection{State abstraction module\label{sub:SAM}}

\begin{figure}[htb]
\centering
\fbox{
\begin{minipage}[t]{\linewidth}
\begin{tabbing}
xxxx \= xxxx \= xxxx \= xxxx \= xxxx \= xxxx \= xxxx \= \kill
$s_p \leftarrow$ primary prototype for $s$ \\
\pushtabs
$s_{p2} \leftarrow$ n\=earest-neighbor prototype for $s$, \\
    \> (if $s_p$ isn't nearest) \\
\poptabs
{\tt investigate}($s$) \\
{\tt investigate}($s_p$) \\
update $Q(j, \cdot)$ by $Q1(s_p, \cdot)$ \\
\\
\pushtabs
if  \=({\tt should\_split}$(s, s_p) =$ no) \\
    \> or ({\tt reliable\_prototype}$(j) =$ no) then \\
\poptabs
    \> update $Q(j, \cdot)$ by $Q1(s, \cdot)$ \\
else split $j:$ \\
    \> reduce reliability info for $j$ \\
    \> if $s_{p2}$ exists then {\tt investigate}($s_{p2}$) \\
\pushtabs
    \> if (\={\tt should\_split}$(s_p, s_{p2}) =$ no) \\
    \>     \>or ({\tt should\_split}$(s, s_{p2}) =$ yes) then \\
\poptabs
    \> \> add $s$ as a new prototype \\
\\
    \> if ({\tt should\_split}$(s_p, s_{p2}) =$ yes) then \\
    \>    \> detach $s_{p2}$ from $s_p$ \\
    \>    \> if we did not add $s$ as prototype \\
    \>    \>    \> update $s_{p2}$ with $Q1(s, \cdot)$ \\
\\
if MERGE\_TIMER() then \\
    \> consolidate compatible states \\
\end{tabbing}
\end{minipage}
}
\caption{\tt update\_representation($s, j$)}
\label{fig:algfe}
\end{figure}

\begin{figure}[htb]
\centering
\fbox{
\begin{minipage}[t]{\linewidth}
\begin{tabbing}
xxxx \= xxxx \= xxxx \= xxxx \= xxxx \= xxxx \= xxxx \= \kill
\pushtabs
if \= (all actions are reliable from $s_1,$ and \\
   \> all actions are reliable from $s_2,$ and \\
   \> $\mbox{\tt compatible}(s_1, s_2) =$ no) \\
\poptabs
then \\
    \> return yes \\
else \\
    \> return no
\end{tabbing}
\end{minipage}
}
\caption{\tt should\_split($s_1, s_2$)}
\label{fig:algsplit}
\end{figure}
Figure~\ref{fig:algfe} outlines the feature extraction algorithm,
implemented by the function {\tt update\_representation($s, j$)}.  The
timer FE\_TIMER() causes this function to be invoked at regular intervals,
allowing time for the action values to settle between changes to the
representation (see Figure~\ref{fig:processstack}).  The function
{\tt reliable\_prototype}($j$) tests whether $Q(j, b)$ has been
updated at least MIN\_UPDATES times for {\em every} action $b;$
this is a stricter test than {\tt reliable\_source($j$),} which guards
the action value update in the top level of the algorithm
(Figure~\ref{fig:algtop}).  The
function {\tt compatible}($s_1, s_2$) applies the state compatibility
criterion given in Definition~\ref{3def:compatibility}, taking the value
of $\delta = \epsilon.$

The basic idea is this:  when investigating surprising states,
if the current state appears to be incompatible with its classification
because of a policy or value difference that results in
significant lost reward, then add the state as the seed of a new
state-space category.  Occasionally, consider whether the
representation may be simplified by merging compatible regions,
and whether merged regions ought to stay merged.

\subsection{Methodology}

How should we test an algorithm for on-line feature extraction?  We want
to know whether the algorithm produces high quality representations for
the agent's task.  A common test is to simply evaluate the performance of
an agent that uses the algorithm to construct its representation for the
task as it goes about learning the task.  Although this technique is
often seen in the literature,  good performance in the task does not
necessarily indicate a high quality representation;  evaluating a
representation is different from simply evaluating performance.

Several other criteria may help evaluate the quality of the
representation.  If the representation contains a small number of
features, that may be taken as evidence of cognitive economy,
provided that those features allow the agent to make the necessary
distinctions in its task.  If the representation is reusable by
other agents, that is evidence that it captures
important features of the task, rather than artifacts of a particular
training regimen.  If the representation allows good performance
from a variety of starting points, that may indicate a high level
of quality throughout the relevant parts of the state-space.
Therefore, our evaluation methodology should test the representation
independently of the system which produced it, and should exercise
the representation over a significant portion of the state-space.
Our quality assessment will be based on the number of features and
the effectiveness of the representation in the task.

The effectiveness of the representation may be shown most effectively
by a learning curve that plots task performance against training
time for learning the action values.  Learning curves are
especially useful for evaluating representations, because a single
measurement of either learning speed or performance is likely to
mislead.  A single measurement of learning speed tends to
favor small representations, which have fewer parameters to be
trained, while a single measurement of performance tends to favor
large representations, which tend to learn more slowly but eventually
produce superior performance.  In addition, learning curves show
whether a representation reliably supports good performance, or
only results in occasional successes.  If a learning curve is
produced, it should be averaged from multiple experimental runs,
in order to minimize system initialization effects.  (Even though the
algorithm may be deterministic, its implementation may rely on
random numbers for action selection when its preferred action
set contains more than one action).

The experiments reported here consisted of two stages: a
representation generation stage and a representation testing stage.
In the generation stage, a learning system applied the algorithm
described above, constructing its representation as it learned
the task.  The output of this stage is the specification of a new
state-space representation for the task.  For the algorithm to
succeed, it must not only learn to perform well in the task,
but must produce a representation that captures the important
features of the problem, in a form reusable by other agents.

The testing stage consisted of inserting the generated representation
into another reinforcement learning system and producing learning
curves for this new system.  Although the representation is now
fixed, the system must still learn its action values from scratch,
which is why it improves with training.  Since the
tester is separate from the agent that produced the representation,
it is able to evaluate different representations fairly, and it
allows us to compare them in a way that controls for the other
aspects of the reinforcement learning problem.  This method of
evaluating learned representations in a separate system appears to
be unique.

To produce a learning curve, the system's action values were reset,
and the system generated a series of learning trials.  The system's
performance was evaluated at the ends of trials, but not after every
trial.  It was important not to interrupt a running trial because the task
studied here has rewards only at the ends of trials.  At the end of a
trial, deciding whether to generate a performance measurement
depended on a two-part test:  the system needed to have completed
either a pre-set number of learning steps or a pre-set number of trials.
This two-part test insured that the system generated enough data
points both early in the run (when trials are short, so the number of
trials dominates) and late (when trials are long, and the number of
learning steps dominates).

Each performance measurement was the median score for a batch
of trials from a set of random starting states.  (For
the puck-on-a-hill task reported here, the scores were just the trial
lengths).
The random starting states were chosen as follows.  A preliminary
series of experiments yielded a set of extreme values for the
state-space coordinates;  the tester's training trials were
started at states from the central third of this observed
state-space, and its test trials were started at states from
a slightly smaller zone---the central quarter of the space.
Initializing trials at random starting states ensures that the representation
is tested over a significant portion of the space, providing a more
meaningful indication of the quality of the representation.

\section{Case Study: Puck-On-A-Hill Task}

The puck-on-a-hill is a bang-bang control task in which rewards
are seen only at the ends of episodes that are normally very long
(up to 5 million steps) and include tight cycles.  Performance in
this task depends on the adequacy of the state-space representation.
A Q-learning agent performs poorly with some seemingly reasonable
representations for this task;  but analysis of the task leads to a
simple, two-category representation that is optimal in a sense
described below.  Therefore, this is an effective demonstration
task for evaluating algorithms for on-line feature extraction.

\begin{figure}[htb]
   \centerline{\includegraphics[width=\linewidth]{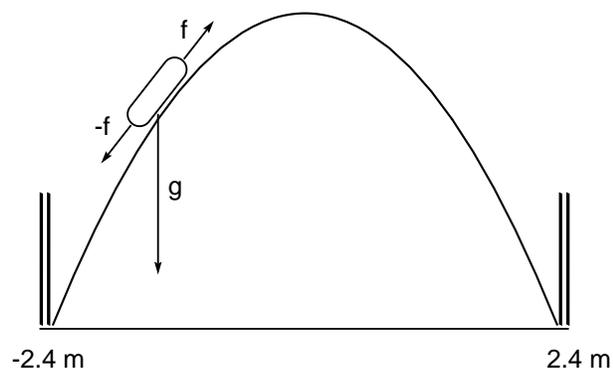}}
   \caption{The puck-on-a-hill task}
   \label{fig:puckhill}
\end{figure}

In the puck-on-a-hill task, the agent controls a puck which it must
learn to push to the left or the right to keep the puck
balanced on the top of a hill.  The agent's only reinforcement
comes when the puck falls too far down the hill on either side and hits
the containing wall.  When that happens, the agent is given a
reward of $-1,$ and the episode ends.  This task has been
studied previously in \cite{Hu96} (although with a slightly different
form for the equations of motion) and in \cite{Finton02}.
Figure~\ref{fig:puckhill} illustrates the task.

The puck-on-a-hill is similar to the familiar pole-balancing task
\cite{MichieChambers68,AndersonMiller90,GevaSitte93} but with a
two-dimensional state-space
having components for position and velocity only.  The simulation
details are as follows.

\begin{tabular}{ll}
\mbox{ }         & \\
State            & $x$, position of puck (meters) \\
                 & $v$, velocity of puck (m / sec) \\
            \\
Control          & $f$, force on puck (Newtons) \\
            \\
Constraints      & $-2.4 < x < 2.4$ \\
                 & $f = \pm 3.0$ \\
            \\
Equation of hill & $y = - \beta x^2$ \\
            \\
Parameters       & $\beta = 0.3$ (hill curvature) \\
                 & $g = 9.8$  m / $\mbox{sec}^2$ (grav. accel.) \\
                 & $m = 1.0$ kg (mass of puck) \\
                 & $\Delta = 0.02$ sec, \\
                 & \mbox{   } (sampling interval) \\
\mbox{ }         &  
\end{tabular}

Positive $x$ represents a position on the right side of the hill.  The
corresponding angle of the cart with the hill is given by $\theta,$ where
positive $\theta$ represents a position on the left side of the hill.  Positive $f$
pushes the puck toward the right.

The equations of motion are as
follows.
\begin{eqnarray*}
    x(t+1) & = & x(t) + \Delta v(t) \\
    v(t+1) & = & v(t) + \Delta \frac{(f(t) - m g \sin \theta(t)) \cos \theta(t)}{m} \\
    \theta(t) & = & \arctan( - 2 \beta x(t) )
\end{eqnarray*}

\subsection{Analysis}

Some analysis of the task will help us understand what makes for a good 
representation.
The puck's acceleration is determined by its thrusters and the
downward pull of gravity.  Near the center of the hill, the thrusters
dominate the force of gravity:  the puck can push itself back
to the crest of the hill as long as its prior velocity is not too
large.  Away from the center, the hill's slope becomes increasingly
steep, so that gravity overwhelms the contribution of the puck's
thrusters.  Therefore, once the puck has fallen too far down the
hill, it loses the ability to climb back up, and fails shortly
after.

Just where this ``point of no return'' lies depends on the puck's
velocity.  From the equations of motion, we see that the acceleration
on the puck is zero when
\[  f(t) = m g \sin \theta(t) \]
\[
    \frac{f}{m g} = \sin \arctan(-2 \beta x)
\]
\[
    \frac{\tan \arcsin \frac{f}{m g}}{-2 \beta} = x
\]
Substituting the values of the parameters,
$f = \pm 3.0 \mbox{ Newtons, } g = 9.8 \mbox{ m} / \mbox{s}^2,
\beta = 0.3, m = 1.0 \mbox{ kg},$ we have
\[
    x = \frac{\tan \arcsin \frac{\pm 3.0}{(1.0)(9.8)}}{(-2)(0.3)} \doteq \pm 0.54
\]
At this point, a puck with zero velocity can hold its position on the hill;
if we place a stationary puck farther out, it cannot avoid falling down the hill.
On the other hand, if the puck is already moving up the hill, its existing
velocity may be sufficient to carry it back into the controllable region,
even starting from a point farther down the hill.  Therefore,
the ``point of no return'' lies farther down the hill with higher puck velocities.

\begin{figure}[htb]
   \centerline{\includegraphics[width=\linewidth]{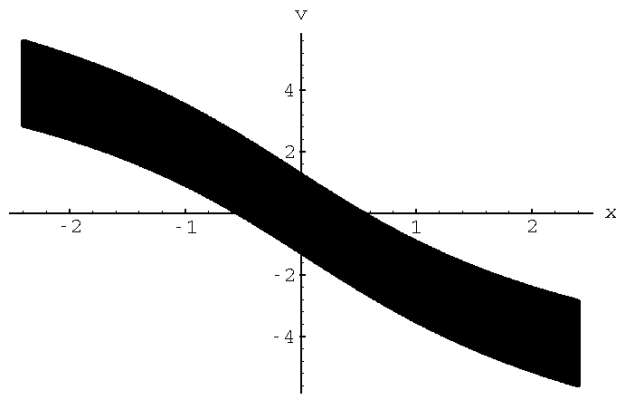}}
   \caption{Controllable states:  states outside this band result in failure.}
   \label{fig:puckswath}
\end{figure}

The agent must keep the puck within the region where its thrusters
are effective in controlling the puck.  We will call these states
{\em controllable states,} and we will call the states that are
past a ``point-of-no-return'' {\em doomed states.}
Figure~\ref{fig:puckswath} shows the controllable states, which
form a band falling roughly diagonally through the middle of
the state-space.  This figure was produced by running puck experiments
at each point of a very fine grid.  (Resolution was 0.01 in both
$x$ and $v$).  The remaining states are all doomed states, from
which the puck cannot avoid falling down the hill and hitting a wall,
no matter what actions it takes.

\begin{figure}[htb]
   \centerline{\includegraphics[width=\linewidth]{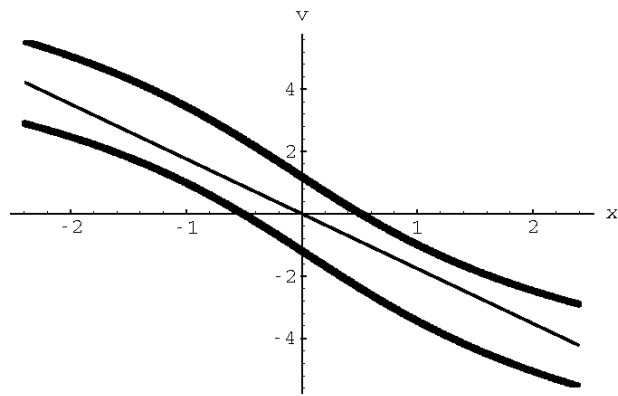}}
   \caption{An ideal representation: must-push-left states (top curve)
                   and must-push-right states (bottom curve) are separated by
                   the diagonal line.}
   \label{fig:puckcritical}
\end{figure}

Since hitting a wall is the only source of reinforcement in this task,
the best possible return is $0,$ obtained by a policy that keeps
the puck within the controllable zone.  Any policy that does so is
therefore an optimal policy for this task.  If pushing right from a controllable
state $s_c$ results in a doomed state, then we may classify $s_c$
as a ``must-push-left'' state.  Similarly, if pushing left results
in a doomed state, we may classify $s_c$ as a ``must-push-right''
state.  If both left and right lead to other controllable states,
we may classify $s_c$ as a ``don't care'' state.  The critical
states are the must-push-right and must-push-left states, where
the agent's next action determines whether it succeeds in the task.
These states are on the edges of the controllable zone;  the states
in the middle of the controllable zone are don't-care states because
neither action will push the puck past the boundary of the zone.
Figure~\ref{fig:puckcritical} shows the critical states: the
must-push-left states make up the top curve, and the must-push-right
states make up the bottom curve.  These plots were determined by
testing each controllable state found in the earlier simulation,
evaluating the controllability of the states which result after a
single push to the left or right.

This analysis shows that simply pushing toward the center of the
hill is not an optimal policy.  For example, if the puck is moving fast
enough toward the right, it may need to push to the left, even when
it is already on the left side of the hill.  Otherwise, it may be unable
to slow down on the other side and avoid hitting the right wall.
Any policy that pushes to the left in the must-push-left states (the
top curve in Figure~\ref{fig:puckcritical}) and pushes to the right in
the must-push-right states (the bottom curve) is an optimal policy.
Therefore, any representation that separates these two classes of
states will be adequate for learning an optimal policy.  For example,
we can simply bisect the controllable zone by the line $v = -1.7615 x,$
(Figure~\ref{fig:puckcritical}).
Since this representation cleanly separates the must-push-right states
from the must-push-left states, it allows the system to learn an optimal
policy.  In addition, this is one of the simplest possible representations
that preserves the necessary distinction.  Therefore, this diagonal-split
representation is a useful benchmark for evaluating other representations.

\subsection{Results}

The results compare the performance of  a test system under different
state-space representations.   Each representation was evaluated by
inserting it into the test system and generating a series of 10
learning curves, which were then averaged.  The learning curves
plot performance against the number of training steps experienced by
the test system.  Each performance score is the median trial length
for a batch of 50 trials conducted with learning turned off.  The
test trials were stopped if they reached 5,000,000 steps.  After
50,000 steps of training, the diagonal-split representation and
the learned representation both attained averaged performance scores
of 5,000,000 steps.

\subsubsection{Generated representations}

\begin{figure}[htb]
   \centerline{\includegraphics[width=\linewidth]{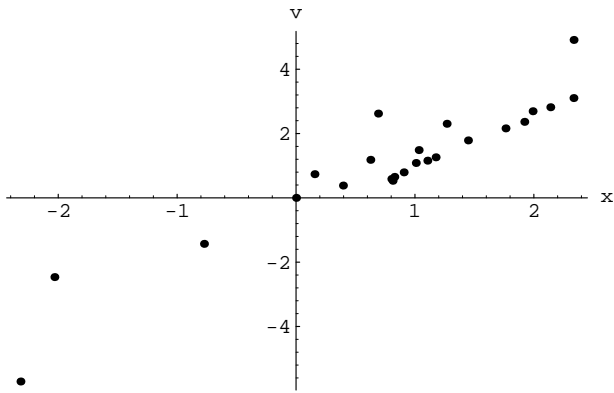}}
   \caption{Representation constructed automatically, from scratch (24 categories).}
   \label{fig:puckrep24}
\end{figure}

Starting from scratch, the system generated a representation consisting
of 24 prototype states, shown in Figure~\ref{fig:puckrep24}.
Although the prototypes have a slightly asymmetric layout, their
placement allows the system to easily identify points that are closer
to the ``must-push-right'' boundary than the ``must-push-left'' boundary,
and {\em vice versa.}  The system adds prototypes from states that it
visits, sometimes in the later stages of a failing trial.  The reason that
the system kept generating points farther out from the center is most
likely that it learned a more negative state value, $V1(s),$ for states
closer to failure points.  Although this value difference turned out to
be unimportant in the puck task, there exist other tasks where it would
have been a critical distinction. 

\begin{figure}[htb]
   \centerline{\includegraphics[width=\linewidth]{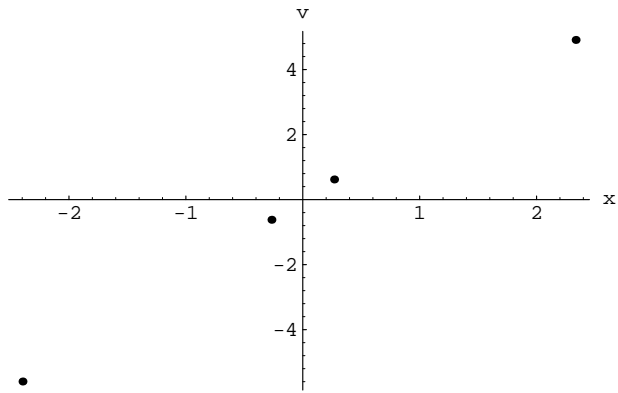}}
   \caption{Representation constructed from a good seed representation.}
   \label{fig:puckrep4}
\end{figure}

In another experiment, the system was seeded with a representation
consisting of the two $(x, v)$-space points  (0.2680, 0.6200) and
(-0.2680, -0.6200).  These points are on either side of the
controllable zone;  although the line connecting them is not quite
perpendicular to the line of the diagonal-split representation
described earlier, these two points were thought to be sufficient
to distinguish must-push-left points from must-push-right points.
The objective of this second experiment was to verify that the
state compatibility criteria do not lead to the generation of
unnecessary states.  This was confirmed by the resulting representation,
which simply added two states at the usual failure points of the task.
Figure~\ref{fig:puckrep4} shows the representation.  The learning
process which produced it required 207 trials, with the last trial
continuing for over 100 million steps.

\subsubsection{Control representations}

The results compare the performance of  the 24-category generated
representation with the performance of four other representations:
the diagonal-split representation described above, a uniform
$10 \times 10$ grid partitioning, a representation inspired by
Variable Resolution Dynamic Programming \cite{Moore91}, and a representation
designed to maintain controllability \cite{Hu96}.

\begin{figure}[htb]
   \centerline{\includegraphics[width=\linewidth]{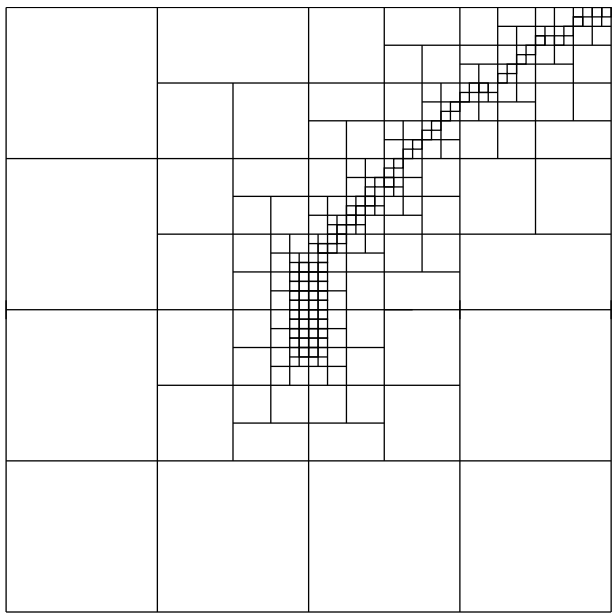}}
   \caption{A representation inspired by Variable Resolution
                   Dynamic Programming.}
   \label{fig:moorepuck}
\end{figure}

Variable Resolution Dynamic Programming (VRDP) produces a partitioning
of the state-space with the highest resolution at states visited
during experimental trials.  Away from these experimental trajectories,
resolution falls off gradually according to a constraint on
neighboring regions.  The experimental trials are
``mental practice sessions'' conducted according to an internal
model being learned by the agent.  For the studies reported here,
the representation was constructed from two trials using the puck
task environment: an initial trial in which the agent always pushed
to the right, and a successful trial in which the agent succeeded
in keeping the puck in the center of the hill for over 100,000
steps.  (The successful trial was taken from a system with the
uniform $10 \times 10$ partitioning of the space).  Because VRDP
initializes the representation to a single box, the initial trial
consisted of selecting the same action repeatedly (since the
policy for all states is the policy of that single box).  When the
representation was fine enough to allow good performance, mental
practice sessions would focus on the states seen in the successful
trial.  Therefore, VRDP would be likely to visit the same points
in mental practice sessions which were visited in the two experimental
trials---and most likely, additional points as the
representation was being learned and performance was still improving.
Therefore, this representation is probably an idealized version of
the application of VRDP to the puck task.  As in \cite{Moore91},
the highest resolution of each state-space coordinate was found by
performing six binary splits of that coordinate.  Taking the
state-space dimensions to be $[-2.4, 2.4] \times [-5.5, 5.5],$ this
resulted in the smallest distinctions being $\Delta x = 4.8 / 64
= 0.075$ and $\Delta v = 11.0 / 64 = 0.171875.$  Figure~\ref{fig:moorepuck}
illustrates the resulting representation.

\begin{figure}[htb]
   \centerline{\includegraphics[width=\linewidth]{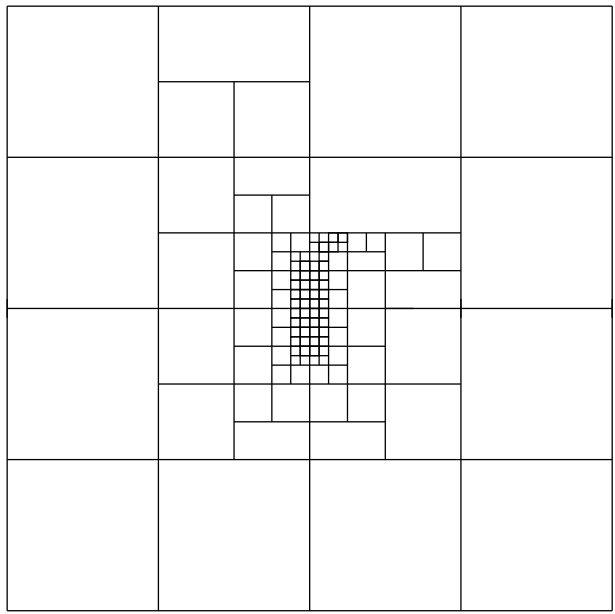}}
   \caption{Enhanced VRDP representation.}
   \label{fig:moorepuck2}
\end{figure}

Unfortunately, this representation performed poorly, attaining a
maximum averaged performance score of 2215 steps.  (Both the
diagonal-split representation and the generated representation
achieved averaged scores of 5,000,000 steps).  One reason for this
poor performance may be that the partitioning is very fine along
the path from the origin to the failure point of the first trial.
As a result, reward from a failure must pass through a very
long series of intermediate boxes before it reaches the critical
states where the agent can actually control the puck.  To test this
explanation, I made a second VRDP-inspired representation, shown
in Figure~\ref{fig:moorepuck2}.  Although this representation does
not entirely observe the constraint on neighboring regions, it
removes most of the boxes resulting from the initial failed trial.
Since this representation performed much better than the original,
it replaces the original VRDP representation in the comparison
plots which follow.

\begin{figure}[htb]
   \centerline{\includegraphics[width=\linewidth]{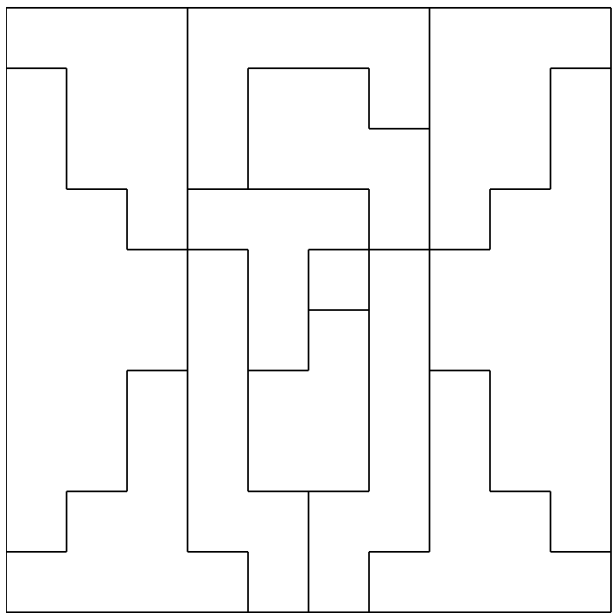}}
   \caption{Representation designed to limit the loss of controllability
                   (from \protect\cite{Hu96}).}
   \label{fig:yendopuck}
\end{figure}

The other representation, shown in Figure~\ref{fig:yendopuck},
was taken from \cite{Hu96}.  This representation attempts to
limit the agent's loss of controllability, according to an off-line
analysis computed in terms of a model of the task:  First,
compute the worst-case deviations between possible
trajectories of the agent that start from different states;  then divide
the state-space into regions small enough that when this
deviation is integrated over all pairs of states in a region, the resulting
controllability error is less than a tolerance.
This representation was part of an Adaptive Heuristic Critic system
\cite{BartoSuttonAnderson83} which learned to balance the
puck for over 10,000 steps, after an average of 13 trials and 2000
training steps.    Since \cite{Hu96}
assumed that trials always start at
(0, 0), the experiments reported here may have been a more severe
test of this representation than the original study, because my test system
starts trials at randomly-chosen starting points.

\subsubsection{Learning curves}

\begin{figure}[htb]
   \centerline{\includegraphics[width=\linewidth]{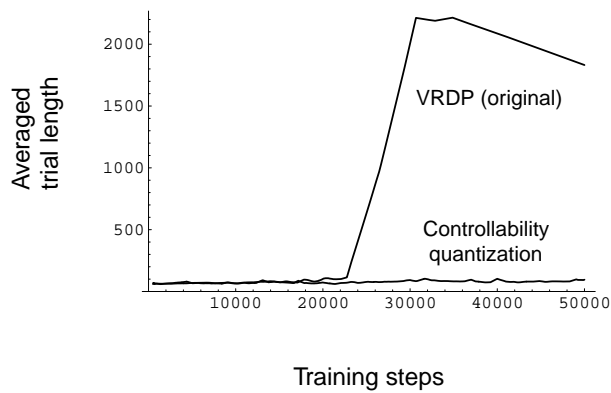}}
   \caption{Averaged performance curves for the original VRDP representation
                    and the controllability quantization.}
   \label{fig:puckcurves}
\end{figure}

\begin{figure}[htb]
   \centerline{\includegraphics[width=\linewidth]{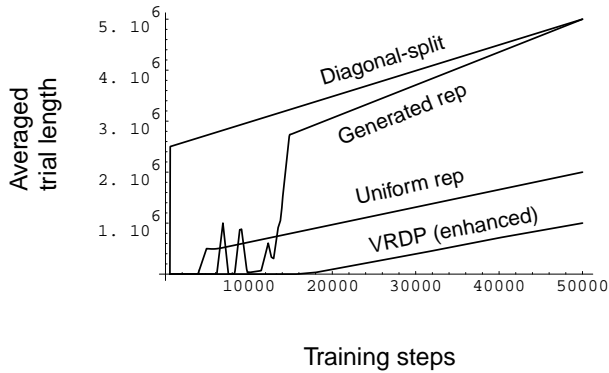}}
   \caption{Averaged performance curves for the four best representations.}
   \label{fig:puckcurves2}
\end{figure}

Figure~\ref{fig:puckcurves} plots the performance for the original
VRDP representation (top curve) and the controllability
quantization (bottom curve).  Note that the performance scores are
all under 2500.  Figure~\ref{fig:puckcurves2} shows the averaged
curves for the remaining representations.  From the top, these are
the diagonal-split representation, the representation generated by
the learning system, the uniform $10 \times 10$ grid, and the
enhanced VRDP representation.  The number of categories for
these representations are, respectively, 2, 24, 100, and 117.

These results illustrate several points.  First, visited or
frequently-seen states are not necessarily important ones.  Second,
irrelevant state-space distinctions can hinder learning, as in the
original VRDP representation.  Third, the important areas of the
space are those where the agent's decision makes a critical difference
in performing the task.  The representations that made the relevant
distinctions (must-push-left versus must-push-right states) in the
simplest way resulted in the most efficient learning.

The cognitive economy approach resulted in a system that was able
to automatically construct a good representation from scratch.
The representation it constructed had a small number of categories
(24), and proved effective in the task.  When given an effective
seed representation, the system made minimal additions, indicating
an ability to discern relevant distinctions.

\section{Discussion and related work}

\subsection{Assumptions}

The system presented in this paper makes several assumptions. 
First, the criteria for policy distinctions assume that the agent's
action is chosen from a discrete set of actions.  Therefore, the
preferred action set is also discrete.  Tasks with continuous-valued
action choices will require criteria that consider
how the {\em range} of action affects the overall reward.

The active system explores its state-space at the ends of trials, which
presupposes that the agent's task is episodic.  Non-episodic tasks
can sometimes be made episodic by choosing certain states as terminal
states;  alternatively, the system could simply conduct its explorations
at regular intervals.

A more serious limitation is that some real-world tasks cannot
allow the controller to reset the system state at will (although
this is no problem for any task which is solved through simulation).
As discussed below, the system can be implemented in non-active
versions that would handle such tasks.

The system also depends on the task not
being too stochastic.  Otherwise, a ``surprising'' criterion would
need to be more sophisticated than the one presented here, taking
into account trends and averages over very many instances.  The
decreasing learning rate used in the active investigations helps
somewhat, since the learning rate $1 / \mbox{Updates}(j, a)$ causes
the updated value to be the average of all the instances seen.
This is important when the state-space regions are too coarse,
since the action values may appear stochastic even in a deterministic
task, simply because they really belong to different kinds of states
which get updated together.

\subsection{Nearest-neighbor representation}

Nearest-neighbor state-abstraction allowed the system to split regions
by simply adding new prototype states.  Like ART
\cite{CarpenterGrossberg88}, this strategy adds a new category for the
current observation if its best classification is a poor match.
Compared to the KD-Tree approach of segmenting the space into
hierarchical boxes \cite{MooreAtkeson95,Reynolds00}, the nearest-neighbor
approach may represent higher-dimensional state-spaces more efficiently
because a few prototypes may still suffice for broad areas of similar
states, instead of needing to populate a set of state-space ``boxes''
whose number grows exponentially with the dimensions of the state-space.
A drawback to the nearest-neighbor approach is that it requires a
sophisticated implementation to work efficiently when the number of
prototypes grows large.

\subsection{Active learning}

The active strategy was a natural choice for generating action-value
profiles for a state, because it provides the values of all actions
from the state at the same time.  This is important because the
values are changing as the agent is learning the task.  Even if
two actions lead to the same resulting state, the agent might
mistakenly believe them to have different values if the values were
computed at different times.  Assessing a state's preferred action
requires knowing the values of all the actions from that state.

The active state investigation strategy is still a form of Q-learning,
since Q-learning does not specify how the value backups must be
distributed among different state-action pairs---only that they continue
to be sampled.  By continuing to push randomly-chosen states onto the
stack, the algorithm ensures that values continue to be sampled.  Given a
static task, a fixed, ``lookup table'' representation, and a learning rate
that decreases appropriately, the convergence guarantees for Q-learning
would therefore extend to the active strategy \cite{WatkinsDayan92}.  (Such
guarantees are much less likely to be found for a system which generates
its representation online, though, which is the main contribution of
this paper).

Active reinforcement learning may learn more efficiently because it
eliminates action value backups away from the ``frontier'' of
learned states;  this is the subject of future research.  This idea
has also been explored by others.  For example, ROUT \cite{BoyanMoore95}
generated Monte Carlo simulations of states on the frontier, and thus
worked backwards from the terminal states to beginning states.  One
significant difference is that ROUT required the task to be acyclic,
and once states were learned, they could not be re-investigated.
The present strategy is more robust, since the use of a replacing
stack allows it to focus on the frontier while reinvestigating states
as needed to learn the representation, or to investigate paths which
prove cyclic (as in the puck task).  Other research has explored the
use of an oracle that provides the results of a state transition
\cite{KearnsMansourNg02}.  Just as in the active strategy presented
here, the system is allowed to choose particular state-action
combinations to investigate;  the oracle simply provides the resulting
state and reward for the transition.

\subsection{Non-active approaches}

There are also non-active strategies for implementing these ideas.
One alternative is to adopt a more stringent test for surprising
states (using {\tt reliable\_prototype()} instead of {\tt
reliable\_source()}, see Section~\ref{sub:SAM});  then immediately
add any state thus selected
as a new prototype for a region, instead of first exploring states
through active investigations.  In the end, one can only decide to
split regions by first making tentative splits and exploring their
effectiveness.  The current system does this when it assesses
the compatibility of a state with its primary prototype,
since it is making a hypothetical separation
between those two states in order to see if the region should be
split.  Splitting on the basis of the surprising test would make a less
tentative initial separation of the states, but the algorithm's
state consolidation procedure could prevent the system from
accumulating unneeded regions.

Another promising approach is to replace the oracle or active
state investigation with an internal model that the agent learns
along with the task.  It can save every state
transition it experiences as an $(s, a, r, s')$ tuple which is
stored with the nearest-neighbor prototype state.  These could be
reorganized as regions are merged or detached, and the tuples could
be reassigned when new prototypes are added.  In this way, the
agent would construct an increasingly accurate model of its world,
and the agent could perform investigations by querying this model
instead of by requesting that the environment reset the system
state for actual trials in the world.  This would also remove the
episodic task limitation, since the agent could conduct investigations
internally, whenever it wished.  This approach has much in common
with Prioritized Sweeping \cite{MooreAtkeson93}, which
uses a priority queue to select states for re-examination: states
whose values have significantly changed, as well as states which
are predicted to lead to them.  The replacing-stack performs a
similar function by giving priority to surprising states,
in order of recency.  The states leading to those states are often
the next states to be considered surprising, leading them to be
investigated as well---unless the change in values did not make a
value or policy distinction.  DYNA \cite{Sutton90} also conducts
internal experiments of this sort, although the state-action pairs
are either chosen randomly or given a priority based on how long
ago they were last updated.  The other difference, of course, is
that DYNA and Prioritized Sweeping assume a fixed representation.

\subsection{State compatibility}

The state compatibility criteria presented in
Definition~\ref{3def:compatibility} are similar to those developed
by others.  For example, \cite{MunosMoore02} acknowledges that ``a
good approximation of the value function at some areas is not needed
if this does not have any impact on the quality of the controller,''
(p. 292).  Their splitting criteria split cells when doing so is
most likely to increase the accuracy of the value function where
there is a transition in the optimal control.  Some of their
state-splitting rules include local criteria for making policy and
value distinctions;  however, their approach targets a slightly
different research problem.  Their method requires an existing
model of the dynamics and reinforcement function for the task,
instead of allowing online learning of an unknown task by experience.
\cite{Reynolds00} presents similar ideas of state-space compatibility,
focusing on decision boundaries and policy distinctions, and
generating a KD-Tree representation of the state-space as the system
learns the task.  Although these compatibility criteria are similar
to the definitions given here, the criteria given in this paper
arise out of an analysis of representational adequacy based on the
idea of incremental regret.  This analysis both links the discussion
to what we know about cognitive economy, and provides an
objective framework for evaluating compatibility criteria.

\section{Conclusions}

Criteria for representational adequacy may indicate relevant training
examples by flagging ``surprising'' states,  leading the agent to
focus its efforts where they will be most useful.  State compatibility
criteria allow the agent to split and merge regions according to
distinctions that are relevant to the task at hand.
Developing representations that focus on relevant distinctions
is one of the abilities needed by reinforcement learning agents
that learn complex tasks in unknown domains.  The criteria presented
here are not {\em ad hoc;}  they are derived from a definition
of learnability that specifies the maximum amount
of lost reward we will accept due to errors of representation.

The criteria for representational adequacy and state compatibility
apply to any general reinforcement learning task in which the agent
learns to predict the long-term reward that results from taking
particular actions from particular states.  
The experimental results indicate that these ideas are useful and
may be applied successfully to real problems.  The system presented
here was able to learn better representations for the puck task than
those supplied by other, well-known methods.  In addition, the ideas
presented here may allow reinforcement learning to scale to more
complex tasks, because they simplify the task in three important ways:
cognitive economy allows the agent to generalize over its state-space
where appropriate, active state investigations allow the agent to
focus on the frontier and avoid useless action-value backups, and the
nearest-neighbor representation allows volumes of state-space with
``smooth'' action values to be represented more sparsely.  These are
promising directions for future research.

%
%

\bibliographystyle{plain}

\bibliography{ref}

\end{document}